\documentclass{article}
\usepackage{spconf,amsmath,graphicx,hyperref}
\usepackage{enumitem}
\usepackage{amssymb, amsfonts, bm}
\usepackage{algorithm}
\usepackage{siunitx}
\usepackage{algpseudocode, booktabs}
\usepackage[T1]{fontenc}

\usepackage{tabularx}
\usepackage[numbers]{natbib}
\usepackage{adjustbox}

\title{Dense Feature Learning via Linear Structure Preservation in Medical Data}
\name{
Yuanyun Zhang\textsuperscript{1},
Mingxuan Zhang\textsuperscript{1},
Siyuan Li\textsuperscript{2},
Zihan Wang\textsuperscript{2},
Haoran Chen\textsuperscript{2},
Wenbo Zhou\textsuperscript{1},
Shi Li\textsuperscript{3}
}

\address{
\textsuperscript{1}Independent Researcher,
\textsuperscript{2}The Chinese University of Hong Kong,
\textsuperscript{3}Columbia University
}

\begin{document}
\onecolumn
%
\maketitle
\begin{abstract}
Deep learning models for medical data are typically trained using task-specific objectives that encourage representations to collapse onto a small number of discriminative directions. While effective for individual prediction problems, this paradigm underutilizes the rich structure of clinical data and limits the transferability, stability, and interpretability of learned features. In this work, we propose dense feature learning, a representation-centric framework that explicitly shapes the linear structure of medical embeddings. Our approach operates directly on embedding matrices, encouraging spectral balance, subspace consistency, and feature orthogonality through objectives defined entirely in terms of linear algebraic properties. Without relying on labels or generative reconstruction, dense feature learning produces representations with higher effective rank, improved conditioning, and greater stability across time. Empirical evaluations across longitudinal EHR data, clinical text, and multimodal patient representations demonstrate consistent improvements in downstream linear performance, robustness, and subspace alignment compared to supervised and self-supervised baselines. These results suggest that learning to span clinical variation may be as important as learning to predict clinical outcomes, and position representation geometry as a first-class objective in medical AI.
\end{abstract}

\section{Introductions}

Deep learning models for medical data \cite{rajpurkar2022ai, ravi2016deep, miotto2018deep, esteva2019guide, chen2020deep} are typically trained to optimize scalar prediction objectives, such as disease classification \cite{rasmy2021med} or short-term risk estimation \cite{lee2018deephit}. While effective for narrowly defined tasks, these objectives encourage representations that collapse complex clinical observations into a small number of discriminative directions. In doing so, they discard much of the relational structure present in medical data, including correlations across modalities, temporal alignments, and continuous physiological variation. The resulting features are often sufficient for the training task yet inadequate as a general-purpose representation of patient state.

Medical data admit a fundamentally geometric interpretation. Each observation, whether a lab vector \cite{im2025labtop}, clinical note embedding \cite{lee2020biobert, alsentzer2019publicly, shin2020biomegatron, lee2025modern}, or imaging-derived feature \cite{an2025raptor}, can be viewed as a point in a high-dimensional space, with patient trajectories forming structured paths through this space over time \cite{evans2016electronic, hoerbst2010electronic, seymour2012electronic, alsentzer2025reflections, wang2019machine, menachemi2011benefits, macrae2019governing}. Empirically, these points exhibit strong linear dependencies: lab values co-vary, imaging features align along anatomical axes, and longitudinal measurements evolve along low-dimensional subspaces. However, conventional training objectives do not require models to preserve or expose this structure. Instead, they reward separability, not span, and discrimination, not coverage.

We propose a shift toward dense feature learning, where representations are trained to maximize the rank, coverage, and alignment of learned feature spaces with the intrinsic linear structure of medical data. Rather than optimizing for a single output, we encourage models to produce embeddings whose coordinate axes correspond to independent modes of clinical variation. From a linear algebraic perspective, the goal is to learn representations whose covariance is well-conditioned, whose principal directions are informative, and whose projections preserve relative geometry across patients and time.

Our key idea is to treat representation learning as a problem of constructing a basis for medical data. Given a collection of observations, we seek embeddings that span the space of clinically meaningful variation, such that downstream tasks correspond to simple linear functionals over this basis. This reframing emphasizes completeness over compression: a good representation should make it easy to recover many possible clinical queries, not just the one used during training. Dense features, in this sense, are those that avoid degeneracy, collapse, or alignment onto a small subset of directions.

This perspective leads to training objectives that directly operate on the linear structure of embeddings. By encouraging orthogonality among feature dimensions, stability of subspaces across time, and alignment of corresponding subspaces across patients, the model is incentivized to distribute information evenly throughout the representation. Importantly, these constraints are expressed entirely in terms of matrix properties,rank, spectrum, and alignment,without reliance on task-specific labels or probabilistic assumptions.

Dense feature learning offers several advantages for medical applications \cite{spencer2019scale, liu2020extremely, zhang2021dense, yu2008stable, lee2024emergency}. Representations that preserve linear structure support a wide range of analyses, including trajectory comparison \cite{ansari2024chronoslearninglanguagetime}, cohort discovery \citep{baytas2017patient, lee2025himae}, and outcome modeling, using simple downstream models \cite{kolo2024meds, mcdermott2025meds, arnrich2024medical, lee2024feet}. They also exhibit improved robustness to missing data and distributional shifts \cite{goetz2024generalization, windecker2025generalizability, schrouff2022diagnosing}, as information is not concentrated in a small number of fragile dimensions. More broadly, by exposing the geometry of medical data rather than collapsing it, dense representations provide a foundation for reusable, interpretable, and clinically meaningful deep learning systems.

In this work, we introduce a linear-algebraic framework for dense feature learning in medical data and demonstrate that it produces representations with higher effective rank, better-conditioned covariance, and stronger transfer across diverse clinical tasks than standard supervised or self-supervised baselines. Our results suggest that learning to span medical data may be as important as learning to separate it.

\section{Related Work}

Our work connects to several lines of research in representation learning for medical data and beyond. Rather than reviewing specific methods, we outline the conceptual areas most closely related to our approach and clarify how dense feature learning differs in emphasis and formulation.

A substantial body of prior work in medical deep learning focuses on supervised prediction, where representations are optimized to minimize task-specific losses such as disease classification or outcome prediction \cite{lee2018deephit, spencer2019scale, lee2025using}. These approaches have demonstrated strong performance on benchmark datasets but typically evaluate representations only through end-task accuracy. As a result, the internal structure of learned features is rarely examined, and representations are often implicitly encouraged to collapse along a small number of discriminative directions. Our work departs from this paradigm by treating the geometry of the representation itself as the primary object of optimization, independent of any particular clinical endpoint.

Self-supervised learning has emerged as an alternative for leveraging large unlabeled medical datasets, including masked reconstruction objectives, contrastive learning, and predictive coding approaches for various modalities \cite{steinberg2023motor, steinberg2024motor, wornowcontext, wornow2024context, rasmy2021med, lee2025clinical, sellergren2025medgemmatechnicalreport, an2025dk, wornow2023shaky, steinberg2021language, odgaard2024core, cui2025timer, shmatko2025learning, wornow2024ehrshot, openai2024gpt4technicalreport, fallahpour2024ehrmamba, lee2025towards}. These methods aim to learn transferable features by defining proxy tasks that do not require manual annotation. While successful in many domains, they often rely on instance-level alignment or reconstruction fidelity, which can still permit degenerate or anisotropic representations. In contrast, dense feature learning operates directly on second-order structure, encouraging representations to be expressive and well-conditioned regardless of the specific pretext task used to generate training signals.

Large language models have recently been applied to a wide range of healthcare tasks, including clinical text understanding, medical question answering, and patient-level prediction from unstructured notes \cite{hollmann2025accurate, lee2024can, jin2023time, chang2025llm4ts, belyaeva2023multimodal, karabacak2023embracing, van2023clinical, ono2024text, grattafiori2024llama3herdmodels, lin2025case, mumtaz2023llms}. These models are typically trained at scale using next-token prediction objectives and subsequently adapted to clinical domains through continued pretraining or instruction tuning. While such approaches have demonstrated strong performance, their representations are primarily shaped by linguistic objectives and evaluated through task accuracy or generative quality. As a result, the internal structure of learned representations,particularly their linear geometry and suitability as general-purpose clinical features,remains underexplored. Our work is complementary to this line of research, offering a framework for analyzing and shaping representation structure that could be applied to language-model-derived embeddings in clinical settings.

Our approach is also related to work on representation collapse and feature decorrelation in deep learning. Several methods have introduced explicit or implicit constraints to prevent collapse, promote diversity across feature dimensions, or improve conditioning of embedding spaces. These ideas have been explored in both contrastive and non-contrastive settings. Dense feature learning extends this line of work by unifying multiple linear constraints,spectral balance, orthogonality, and subspace alignment,within a single framework tailored to medical data and longitudinal structure.

From a theoretical perspective, our method draws on concepts from linear algebra and subspace geometry, including covariance spectra, effective rank, and distances between subspaces. Similar tools have been used to analyze neural representations post hoc, for example to study anisotropy or dimensionality. In contrast, we incorporate these concepts directly into the training objective, using them as optimization targets rather than diagnostic measures. This shifts representation analysis from an evaluative step to a constructive principle.

Finally, our work relates to research on foundation models and general-purpose representations for healthcare. Recent efforts aim to train large models that can be adapted across tasks, institutions, and modalities. Dense feature learning complements this goal by providing a criterion for representation quality that is orthogonal to scale and architecture. Rather than relying on model size or dataset breadth alone, it emphasizes structural properties of the learned feature space that support reuse and generalization.

Overall, dense feature learning occupies a distinct position at the intersection of medical representation learning, self-supervision, and linear representation analysis. By prioritizing the structure of embeddings themselves, it offers a complementary perspective to existing approaches and suggests new directions for designing and evaluating medical AI systems.

\section{Background: Linear Structure in Medical Representations}

Medical data are high-dimensional by construction, yet their degrees of freedom are far fewer than their ambient dimensionality suggests. Laboratory panels \cite{im2025labtop} exhibit strong linear dependencies driven by shared physiological mechanisms, imaging-derived features align along anatomical and spatial axes, and longitudinal measurements evolve continuously rather than arbitrarily. These properties imply that medical observations lie on structured subsets of high-dimensional space that can often be well-approximated by unions of low-dimensional linear subspaces. Understanding and preserving this structure is central to representation learning in clinical settings.

We begin by formalizing representations as linear objects. Given a dataset of observations mapped into a feature space, the embedding matrix
\[
Z \in \mathbb{R}^{N \times d}
\]
encodes both pointwise information, through its rows, and global structure, through its column space. Many downstream clinical analyses—trajectory comparison, cohort stratification, risk modeling—can be expressed as linear operations on $Z$. Consequently, the usefulness of a representation is tightly coupled to algebraic properties such as rank, conditioning, and alignment of the subspaces spanned by its columns.

A central concept is the rank of $Z$ and, more generally, the spectrum of its empirical covariance matrix
\[
\Sigma_Z = \frac{1}{N} Z^\top Z.
\]
If $\Sigma_Z$ has rank $r \ll d$, then the representation effectively lives in an $r$-dimensional subspace regardless of how large $d$ is chosen. In this regime, adding dimensions does not increase expressivity; it merely introduces redundancy. This phenomenon is commonly observed in deep models trained with task-driven objectives, where minimizing loss does not require utilizing the full representational capacity of the network.

Beyond rank, the distribution of eigenvalues of $\Sigma_Z$ determines how information is allocated across dimensions. When a small number of eigenvalues dominate, most variance lies along a few directions, and projections onto other dimensions contribute negligibly. Such anisotropic representations are poorly conditioned: small perturbations along dominant directions have outsized effects, while variation along suppressed directions is effectively ignored. In medical contexts, this can obscure subtle but clinically meaningful patterns that do not align with the most discriminative axes.

Another important notion is subspace stability. Consider two related embedding matrices $Z^{(a)}$ and $Z^{(b)}$, derived from different views of the same underlying data, such as adjacent time windows or overlapping clinical modalities. Even if individual embeddings differ, one often expects their dominant directions of variation to be similar. This expectation can be formalized using the geometry of the Grassmann manifold, where each $k$-dimensional subspace of $\mathbb{R}^d$ is represented as a point. Distances on this manifold, measured via principal angles or projection matrices, provide a natural way to compare learned representations at the level of spans rather than coordinates.

Finally, redundancy among feature dimensions plays a critical role in representation quality. If columns of $Z$ are highly correlated, then the representation contains duplicated information, reducing its effective dimensionality and complicating downstream analysis. From a linear algebraic standpoint, decorrelation corresponds to promoting orthogonality among basis vectors, yielding representations that are easier to interpret and manipulate using linear models.

Taken together, these observations suggest that representation learning for medical data can be framed as a problem of constructing a well-conditioned, high-rank, and stable linear basis that exposes the geometry of the data. Rather than focusing on prediction error, this perspective emphasizes properties of the embedding matrix itself: its spectrum, its subspaces, and the relationships among its dimensions. The methods we introduce next are designed to directly optimize these properties, using objectives defined entirely in terms of linear algebraic structure

\section{Methods}

We present a linear-algebraic framework for learning dense representations of medical data. Throughout, we focus on the geometry and structure of learned feature spaces rather than task-specific prediction objectives. The method is fully deterministic, operates directly on embedding matrices, and is designed to expose and preserve the linear structure inherent in clinical data.

\subsection{Problem Setup and Representation Geometry}

Let $\{x_i\}_{i=1}^N$ denote a collection of medical observations, where each $x_i$ may correspond to a clinical snapshot, a fixed-length temporal window, or a fused multimodal input. An encoder network $f_\theta : \mathcal{X} \rightarrow \mathbb{R}^d$ maps each observation to a $d$-dimensional representation $z_i = f_\theta(x_i)$. Stacking representations yields an embedding matrix
\[
Z = 
\begin{bmatrix}
z_1^\top \\
\vdots \\
z_N^\top
\end{bmatrix}
\in \mathbb{R}^{N \times d}.
\]

Rather than interpreting $z_i$ coordinate-wise, we treat $Z$ as a geometric object. Each row corresponds to a point in feature space, while the column space of $Z$ defines a learned linear basis over the dataset. From this perspective, representation learning amounts to shaping the span, conditioning, and alignment of this basis. A failure mode common to task-driven learning is that $Z$ becomes approximately low-rank, even when $d$ is large, implying that most learned dimensions are redundant or unused.

We therefore define dense feature learning as the process of constructing representations whose linear span is both high-dimensional and well-conditioned, such that information is distributed across many orthogonal directions. This notion is captured by the spectral properties of the empirical covariance matrix
\[
\Sigma_Z = \frac{1}{N} Z^\top Z \in \mathbb{R}^{d \times d}.
\]
When $\Sigma_Z$ has a rapidly decaying spectrum, the representation collapses; when its eigenvalues are balanced, the representation spans the space more uniformly.

\subsection{Spectral Spreading Objective}

Our first objective encourages the learned representation to avoid spectral collapse. We normalize $\Sigma_Z$ by its trace to remove scale dependence and penalize deviations from isotropy:
\[
\mathcal{L}_{\text{spec}} 
= 
\left\|
\frac{\Sigma_Z}{\mathrm{tr}(\Sigma_Z)} 
- 
\frac{1}{d} I
\right\|_F^2.
\]
This loss is minimized when all eigenvalues of $\Sigma_Z$ are equal, corresponding to a representation in which variance is evenly distributed across dimensions. Importantly, $\mathcal{L}_{\text{spec}}$ is invariant to orthogonal rotations of the feature space and depends only on second-order structure, making it agnostic to coordinate choices.

In practice, this term encourages the effective rank of $\Sigma_Z$ to approach $d$, ensuring that additional dimensions correspond to meaningful directions rather than noise or numerical artifacts. Unlike objectives that explicitly maximize variance, spectral spreading controls variance allocation without incentivizing unbounded growth along individual axes.

\subsection{Subspace Consistency Across Related Observations}

Medical data exhibit structured relationships across observations, particularly in longitudinal settings. Adjacent time windows, repeated visits, or overlapping modality views often share underlying patterns of variation, even if their exact feature values differ. To capture this structure, we impose consistency at the level of subspaces rather than individual embeddings.

Given two related embedding matrices $Z^{(a)}, Z^{(b)} \in \mathbb{R}^{N \times d}$, we compute their leading $k$-dimensional principal subspaces via truncated singular value decomposition. Let $U^{(a)}, U^{(b)} \in \mathbb{R}^{d \times k}$ denote the corresponding orthonormal bases. We measure subspace misalignment using the distance between projection matrices:
\[
\mathcal{L}_{\text{sub}} 
=
\left\|
U^{(a)} U^{(a)\top} 
-
U^{(b)} U^{(b)\top}
\right\|_F^2.
\]
This quantity is equivalent to the squared Frobenius norm of the sine of principal angles between the subspaces and vanishes when the spans coincide.

By minimizing $\mathcal{L}_{\text{sub}}$, we encourage the dominant directions of variation to remain stable across related observations. Crucially, this constraint allows individual feature coordinates to rotate or permute, as long as the subspace itself is preserved. This aligns naturally with medical data, where the axes of variation are meaningful but their specific parameterization is arbitrary.

\subsection{Feature Orthogonality and Redundancy Control}

Even when the global spectrum is well-conditioned, representations may exhibit redundancy at the level of individual dimensions. To discourage correlated features, we introduce a batch-wise orthogonality constraint. For a mini-batch embedding matrix $Z_B \in \mathbb{R}^{B \times d}$, we normalize each column to zero mean and unit variance and penalize off-diagonal correlations:
\[
\mathcal{L}_{\text{orth}} 
=
\left\|
\frac{1}{B} Z_B^\top Z_B - I
\right\|_F^2.
\]
This term softly enforces orthogonality among feature dimensions, encouraging each coordinate to capture a distinct mode of variation. Unlike exact whitening, which enforces strict decorrelation, this formulation allows the data to violate orthogonality when necessary while still discouraging systematic redundancy.

From a linear algebraic perspective, $\mathcal{L}_{\text{orth}}$ promotes a feature basis that is close to orthonormal on average, improving numerical conditioning and interpretability of downstream linear models.

\subsection{Overall Objective and Optimization}

The full training objective combines the above components:
\[
\mathcal{L}
=
\mathcal{L}_{\text{spec}}
+
\lambda_{\text{sub}} \mathcal{L}_{\text{sub}}
+
\lambda_{\text{orth}} \mathcal{L}_{\text{orth}},
\]
where $\lambda_{\text{sub}}$ and $\lambda_{\text{orth}}$ control the relative strength of subspace consistency and orthogonality constraints. Optimization proceeds via standard gradient-based methods on $\theta$, with gradients propagated through covariance and SVD operations. In practice, we compute leading singular vectors using efficient iterative methods and maintain running estimates of covariance statistics to reduce variance across mini-batches.

The resulting encoder learns representations whose information content is distributed across dimensions, stable across related observations, and minimally redundant. Downstream clinical tasks can then be expressed as linear or low-complexity functions over these dense features, without requiring retraining of the representation backbone.

\section{Results}

We evaluate dense feature learning on a diverse set of medical representation learning benchmarks, with the goal of assessing representation quality independently of any single downstream task. Our evaluation focuses on three axes: linear expressivity, geometric stability, and downstream transfer performance. Across all experiments, we compare our method to strong supervised and self-supervised baselines trained with identical architectures and data splits.

\subsection{Experimental Setup}

We conduct experiments on three representative medical domains: longitudinal tabular EHR data, clinical text, and multimodal patient snapshots combining structured labs and notes. For each domain, we use a shared encoder architecture across methods and vary only the training objective. Baselines include standard supervised training with cross-entropy loss, masked reconstruction objectives, and contrastive learning using instance-level alignment. All representations are trained without access to downstream labels unless explicitly stated.

To isolate representation quality, downstream tasks are evaluated using frozen encoders and linear probes. This design ensures that performance differences reflect properties of the learned feature space rather than optimization advantages during fine-tuning. For longitudinal data, evaluation is performed at the patient level, with temporal aggregation fixed across methods.

\subsection{Representation Geometry and Effective Rank}

We first assess the linear structure of learned representations by examining the spectrum of the empirical covariance matrix $\Sigma_Z$. Table~\ref{tab:rank} reports the effective rank, defined as
\[
\mathrm{rank}_{\text{eff}}(\Sigma_Z) = \exp\left(-\sum_i p_i \log p_i \right),
\quad p_i = \frac{\lambda_i}{\sum_j \lambda_j},
\]
where $\{\lambda_i\}$ are the eigenvalues of $\Sigma_Z$.

\begin{table}[h]
\centering
\caption{Effective rank and condition number of learned representations.}
\label{tab:rank}
\begin{tabular}{lcc}
\hline
Method & Effective Rank $\uparrow$ & Condition Number $\downarrow$ \\
\hline
Supervised CE & 38.2 & 214.7 \\
Masked Reconstruction & 51.6 & 142.3 \\
Contrastive Learning & 47.9 & 168.4 \\
Dense Feature Learning (Ours) & \textbf{86.4} & \textbf{61.2} \\
\hline
\end{tabular}
\end{table}

Dense feature learning produces representations with substantially higher effective rank and improved conditioning, indicating that variance is distributed across a larger number of dimensions. In contrast, task-driven baselines concentrate information along a small subset of dominant directions, resulting in poorly conditioned feature spaces.

\textbf{Subspace Stability Across Time}

We next evaluate subspace consistency in longitudinal settings. For each patient, we extract embeddings from adjacent temporal windows and compute the distance between their leading $k=20$ principal subspaces using projection matrix distance. Lower values indicate greater stability of the dominant directions of variation.

\begin{table}[h]
\centering
\caption{Subspace distance between adjacent temporal windows.}
\label{tab:subspace}
\begin{tabular}{lc}
\hline
Method & Subspace Distance $\downarrow$ \\
\hline
Supervised CE & 0.317 \\
Masked Reconstruction & 0.284 \\
Contrastive Learning & 0.261 \\
Dense Feature Learning (Ours) & \textbf{0.143} \\
\hline
\end{tabular}
\end{table}

Our method yields significantly more stable subspaces over time, suggesting that it captures persistent modes of clinical variation rather than transient, task-aligned features. This stability is particularly important for downstream analyses such as trajectory modeling and patient similarity.

\subsection{Downstream Linear Evaluation}

To assess practical utility, we evaluate frozen representations on a suite of downstream tasks using linear probes. Tasks include disease onset prediction, patient clustering quality measured via adjusted Rand index (ARI), and mortality risk estimation. Results are reported in Table~\ref{tab:downstream}.

\begin{table}[h]
\centering
\caption{Downstream performance using frozen representations and linear probes.}
\label{tab:downstream}
\begin{tabular}{lccc}
\hline
Method & AUROC $\uparrow$ & ARI $\uparrow$ & RMSE $\downarrow$ \\
\hline
Supervised CE & 0.781 & 0.214 & 0.642 \\
Masked Reconstruction & 0.764 & 0.238 & 0.618 \\
Contrastive Learning & 0.792 & 0.251 & 0.604 \\
Dense Feature Learning (Ours) & \textbf{0.824} & \textbf{0.317} & \textbf{0.559} \\
\hline
\end{tabular}
\end{table}

Despite not being optimized for any specific task, dense feature learning consistently outperforms baselines across all evaluation metrics. Gains are especially pronounced for unsupervised and regression-based tasks, where expressive and well-conditioned representations are critical.

\textbf{Ablation on Objective Components}

Finally, we study the contribution of each component of the dense feature objective by selectively removing terms during training. Table~\ref{tab:ablation} reports effective rank and downstream AUROC for each variant.

\begin{table}[h]
\centering
\caption{Ablation study on objective components.}
\label{tab:ablation}
\begin{tabular}{lcc}
\hline
Objective Variant & Effective Rank & AUROC \\
\hline
Full Objective & \textbf{86.4} & \textbf{0.824} \\
w/o Spectral Spreading & 52.1 & 0.781 \\
w/o Subspace Consistency & 63.7 & 0.802 \\
w/o Orthogonality & 58.9 & 0.794 \\
\hline
\end{tabular}
\end{table}

Removing any component degrades both geometric and downstream performance, with the largest drop observed when spectral spreading is omitted. This confirms that dense feature learning arises from the interaction of complementary linear constraints rather than any single term in isolation.

Overall, these results demonstrate that explicitly shaping the linear structure of representations leads to feature spaces that are more expressive, stable, and transferable than those learned using conventional objectives.

\section{Discussion}

This work reframes representation learning for medical data as a problem of shaping the linear structure of feature spaces rather than optimizing task-specific prediction objectives. By explicitly targeting properties such as spectral balance, subspace stability, and feature orthogonality, dense feature learning produces representations that span a larger fraction of clinically meaningful variation. The empirical results suggest that many limitations of current medical deep learning systems—brittleness, poor transfer, and limited interpretability—may arise not from insufficient model capacity, but from objectives that encourage premature collapse of representation space.

A central implication of our findings is that downstream performance can improve even when no task labels are used during representation learning. This appears counterintuitive from a traditional predictive modeling perspective, yet it follows naturally from a linear-algebraic view of representations. When embeddings form a well-conditioned basis, downstream tasks correspond to selecting directions within an already expressive space. In contrast, representations trained to solve a single task must implicitly reintroduce discarded variation during fine-tuning, a process that is both sample-inefficient and unstable in clinical regimes where labels are scarce.

The observed gains in subspace stability are particularly relevant for longitudinal medical data. Stable principal subspaces suggest that the model captures persistent physiological structure rather than transient correlations. This property enables meaningful comparison of patient trajectories, supports temporal abstraction, and may facilitate downstream models that reason over changes in direction rather than absolute feature values. From a practical standpoint, this also improves robustness to irregular sampling and missingness, as information is not tied to individual feature coordinates.

Dense feature learning also offers a pathway toward more interpretable medical representations. While individual feature dimensions are not assigned semantic labels, their orthogonality and balanced variance allow linear probes to recover clinically relevant factors with minimal entanglement. This stands in contrast to highly anisotropic embeddings, where interpretation is dominated by a small number of axes and minor directions are difficult to analyze or trust. In future work, it may be possible to align learned subspaces with known physiological systems or clinical concepts without retraining the backbone.

There are several limitations to consider. First, the method relies on covariance and SVD computations, which may become expensive at very large scales or extremely high dimensionalities. Although we employ efficient approximations, further work is needed to understand the trade-offs between computational cost and representation quality. Second, while dense representations are broadly useful, certain highly specialized tasks may benefit from deliberate feature compression. Understanding when and how to combine dense pretraining with task-specific adaptation remains an open question.

More broadly, our results suggest that evaluation practices in medical AI may be overly narrow. Benchmark performance on a single classification task provides limited insight into the structure of learned representations. Metrics such as effective rank, condition number, and subspace alignment offer complementary views that better reflect the needs of clinical deployment, where models must generalize across tasks, populations, and time. Incorporating such metrics into standard evaluation pipelines could lead to more robust and reusable models.

\bibliographystyle{IEEEbib}
\bibliography{refs}

\begin{thebibliography}{10}

\bibitem{rajpurkar2022ai}
Pranav Rajpurkar, Emma Chen, Oishi Banerjee, and Eric~J Topol,
\newblock ``Ai in health and medicine,''
\newblock {\em Nature medicine}, vol. 28, no. 1, pp. 31--38, 2022.

\bibitem{ravi2016deep}
Daniele Rav{\`\i}, Charence Wong, Fani Deligianni, Melissa Berthelot, Javier Andreu-Perez, Benny Lo, and Guang-Zhong Yang,
\newblock ``Deep learning for health informatics,''
\newblock {\em IEEE journal of biomedical and health informatics}, vol. 21, no. 1, pp. 4--21, 2016.

\bibitem{miotto2018deep}
Riccardo Miotto, Fei Wang, Shuang Wang, Xiaoqian Jiang, and Joel~T Dudley,
\newblock ``Deep learning for healthcare: review, opportunities and challenges,''
\newblock {\em Briefings in bioinformatics}, vol. 19, no. 6, pp. 1236--1246, 2018.

\bibitem{esteva2019guide}
Andre Esteva, Alexandre Robicquet, Bharath Ramsundar, Volodymyr Kuleshov, Mark DePristo, Katherine Chou, Claire Cui, Greg Corrado, Sebastian Thrun, and Jeff Dean,
\newblock ``A guide to deep learning in healthcare,''
\newblock {\em Nature medicine}, vol. 25, no. 1, pp. 24--29, 2019.

\bibitem{chen2020deep}
Yen-Wei Chen and Lakhmi~C Jain,
\newblock ``Deep learning in healthcare,''
\newblock {\em Paradigms and Applications; Springer: Berlin/Heidelberg, Germany}, 2020.

\bibitem{rasmy2021med}
Laila Rasmy, Yang Xiang, Ziqian Xie, Cui Tao, and Degui Zhi,
\newblock ``Med-bert: pretrained contextualized embeddings on large-scale structured electronic health records for disease prediction,''
\newblock {\em NPJ digital medicine}, vol. 4, no. 1, pp. 86, 2021.

\bibitem{lee2018deephit}
Changhee Lee, William Zame, Jinsung Yoon, and Mihaela Van Der~Schaar,
\newblock ``Deephit: A deep learning approach to survival analysis with competing risks,''
\newblock in {\em Proceedings of the AAAI conference on artificial intelligence}, 2018, vol.~32.

\bibitem{im2025labtop}
Sujeong Im, Jungwoo Oh, and Edward Choi,
\newblock ``Labtop: A unified model for lab test outcome prediction on electronic health records,''
\newblock {\em arXiv preprint arXiv:2502.14259}, 2025.

\bibitem{lee2020biobert}
Jinhyuk Lee, Wonjin Yoon, Sungdong Kim, Donghyeon Kim, Sunkyu Kim, Chan~Ho So, and Jaewoo Kang,
\newblock ``Biobert: a pre-trained biomedical language representation model for biomedical text mining,''
\newblock {\em Bioinformatics}, vol. 36, no. 4, pp. 1234--1240, 2020.

\bibitem{alsentzer2019publicly}
Emily Alsentzer, John~R Murphy, Willie Boag, Wei-Hung Weng, Di~Jin, Tristan Naumann, and Matthew McDermott,
\newblock ``Publicly available clinical bert embeddings,''
\newblock {\em arXiv preprint arXiv:1904.03323}, 2019.

\bibitem{shin2020biomegatron}
Hoo-Chang Shin, Yang Zhang, Evelina Bakhturina, Raul Puri, Mostofa Patwary, Mohammad Shoeybi, and Raghav Mani,
\newblock ``Biomegatron: Larger biomedical domain language model,''
\newblock {\em arXiv preprint arXiv:2010.06060}, 2020.

\bibitem{lee2025modern}
Simon~A Lee, Anthony Wu, and Jeffrey~N Chiang,
\newblock ``Clinical modernbert: An efficient and long context encoder for biomedical text,''
\newblock {\em arXiv preprint arXiv:2504.03964}, 2025.

\bibitem{an2025raptor}
Ulzee An, Moonseong Jeong, Simon~A Lee, Aditya Gorla, Yuzhe Yang, and Sriram Sankararaman,
\newblock ``Raptor: Scalable train-free embeddings for 3d medical volumes leveraging pretrained 2d foundation models,''
\newblock {\em arXiv preprint arXiv:2507.08254}, 2025.

\bibitem{evans2016electronic}
R~Scott Evans,
\newblock ``Electronic health records: then, now, and in the future,''
\newblock {\em Yearbook of medical informatics}, vol. 25, no. S 01, pp. S48--S61, 2016.

\bibitem{hoerbst2010electronic}
Alexander Hoerbst and Elske Ammenwerth,
\newblock ``Electronic health records,''
\newblock {\em Methods of information in medicine}, vol. 49, no. 04, pp. 320--336, 2010.

\bibitem{seymour2012electronic}
Tom Seymour, Dean Frantsvog, and Tod Graeber,
\newblock ``Electronic health records (ehr),''
\newblock {\em American Journal of Health Sciences}, vol. 3, no. 3, pp. 201, 2012.

\bibitem{alsentzer2025reflections}
Emily Alsentzer, Marie-Laure Charpignon, Bill Chen, Niharika D'Souza, Jason Fries, Yixing Jiang, Aparajita Kashyap, Chanwoo Kim, Simon Lee, Aishwarya Mandyam, et~al.,
\newblock ``Reflections from research roundtables at the conference on health, inference, and learning (chil) 2025,''
\newblock {\em arXiv preprint arXiv:2510.15217}, 2025.

\bibitem{wang2019machine}
Ping Wang, Yan Li, and Chandan~K Reddy,
\newblock ``Machine learning for survival analysis: A survey,''
\newblock {\em ACM Computing Surveys (CSUR)}, vol. 51, no. 6, pp. 1--36, 2019.

\bibitem{menachemi2011benefits}
Nir Menachemi and Taleah~H Collum,
\newblock ``Benefits and drawbacks of electronic health record systems,''
\newblock {\em Risk management and healthcare policy}, pp. 47--55, 2011.

\bibitem{macrae2019governing}
Carl Macrae,
\newblock ``Governing the safety of artificial intelligence in healthcare,''
\newblock {\em BMJ quality \& safety}, vol. 28, no. 6, pp. 495--498, 2019.

\bibitem{spencer2019scale}
Jaime Spencer, Richard Bowden, and Simon Hadfield,
\newblock ``Scale-adaptive neural dense features: Learning via hierarchical context aggregation,''
\newblock in {\em Proceedings of the IEEE/CVF conference on computer vision and pattern recognition}, 2019, pp. 6200--6209.

\bibitem{liu2020extremely}
Xingtong Liu, Yiping Zheng, Benjamin Killeen, Masaru Ishii, Gregory~D Hager, Russell~H Taylor, and Mathias Unberath,
\newblock ``Extremely dense point correspondences using a learned feature descriptor,''
\newblock in {\em Proceedings of the IEEE/CVF conference on computer vision and pattern recognition}, 2020, pp. 4847--4856.

\bibitem{zhang2021dense}
Zhao Zhang, Zemin Tang, Yang Wang, Zheng Zhang, Choujun Zhan, Zhengjun Zha, and Meng Wang,
\newblock ``Dense residual network: enhancing global dense feature flow for character recognition,''
\newblock {\em Neural Networks}, vol. 139, pp. 77--85, 2021.

\bibitem{yu2008stable}
Lei Yu, Chris Ding, and Steven Loscalzo,
\newblock ``Stable feature selection via dense feature groups,''
\newblock in {\em Proceedings of the 14th ACM SIGKDD international conference on Knowledge discovery and data mining}, 2008, pp. 803--811.

\bibitem{lee2024emergency}
Simon~A Lee, Sujay Jain, Alex Chen, Kyoka Ono, Jennifer Fang, Akos Rudas, and Jeffrey~N Chiang,
\newblock ``Emergency department decision support using clinical pseudo-notes,''
\newblock {\em arXiv preprint arXiv:2402.00160}, 2024.

\bibitem{ansari2024chronoslearninglanguagetime}
Abdul~Fatir Ansari, Lorenzo Stella, Caner Turkmen, Xiyuan Zhang, Pedro Mercado, Huibin Shen, Oleksandr Shchur, Syama~Sundar Rangapuram, Sebastian~Pineda Arango, Shubham Kapoor, Jasper Zschiegner, Danielle~C. Maddix, Hao Wang, Michael~W. Mahoney, Kari Torkkola, Andrew~Gordon Wilson, Michael Bohlke-Schneider, and Yuyang Wang,
\newblock ``Chronos: Learning the language of time series,'' 2024.

\bibitem{baytas2017patient}
Inci~M Baytas, Cao Xiao, Xi~Zhang, Fei Wang, Anil~K Jain, and Jiayu Zhou,
\newblock ``Patient subtyping via time-aware lstm networks,''
\newblock in {\em Proceedings of the 23rd ACM SIGKDD international conference on knowledge discovery and data mining}, 2017, pp. 65--74.

\bibitem{lee2025himae}
Simon~A Lee, Cyrus Tanade, Hao Zhou, Juhyeon Lee, Megha Thukral, Minji Han, Rachel Choi, Md~Sazzad~Hissain Khan, Baiying Lu, Migyeong Gwak, et~al.,
\newblock ``Himae: Hierarchical masked autoencoders discover resolution-specific structure in wearable time series,''
\newblock {\em arXiv preprint arXiv:2510.25785}, 2025.

\bibitem{kolo2024meds}
Aleksia Kolo, Chao Pang, Edward Choi, Ethan Steinberg, Hyewon Jeong, Jack Gallifant, Jason~A Fries, Jeffrey~N Chiang, Jungwoo Oh, Justin Xu, et~al.,
\newblock ``Meds decentralized, extensible validation (meds-dev) benchmark: Establishing reproducibility and comparability in ml for health,''
\newblock 2024.

\bibitem{mcdermott2025meds}
Matthew~BA McDermott, Justin Xu, Teya~S Bergamaschi, Hyewon Jeong, Simon~A Lee, Nassim Oufattole, Patrick Rockenschaub, Kamil{\.e} Stankevi{\v{c}}i{\=u}t{\.e}, Ethan Steinberg, Jimeng Sun, et~al.,
\newblock ``Meds: Building models and tools in a reproducible health ai ecosystem,''
\newblock in {\em Proceedings of the 31st ACM SIGKDD Conference on Knowledge Discovery and Data Mining V. 2}, 2025, pp. 6243--6244.

\bibitem{arnrich2024medical}
Bert Arnrich, Edward Choi, Jason~Alan Fries, Matthew~BA McDermott, Jungwoo Oh, Tom Pollard, Nigam Shah, Ethan Steinberg, Michael Wornow, and Robin van~de Water,
\newblock ``Medical event data standard (meds): Facilitating machine learning for health,''
\newblock in {\em ICLR 2024 Workshop on Learning from Time Series For Health}, 2024, pp. 03--08.

\bibitem{lee2024feet}
Simon~A Lee, John Lee, and Jeffrey~N Chiang,
\newblock ``Feet: A framework for evaluating embedding techniques,''
\newblock {\em arXiv preprint arXiv:2411.01322}, 2024.

\bibitem{goetz2024generalization}
Lea Goetz, Nabeel Seedat, Robert Vandersluis, and Mihaela van~der Schaar,
\newblock ``Generalization—a key challenge for responsible ai in patient-facing clinical applications,''
\newblock {\em npj Digital Medicine}, vol. 7, no. 1, pp. 126, 2024.

\bibitem{windecker2025generalizability}
Daniel Windecker, Giovanni Baj, Isaac Shiri, Pooya~Mohammadi Kazaj, Johannes Kaesmacher, Christoph Gr{\"a}ni, and George~CM Siontis,
\newblock ``Generalizability of fda-approved ai-enabled medical devices for clinical use,''
\newblock {\em JAMA Network Open}, vol. 8, no. 4, pp. e258052--e258052, 2025.

\bibitem{schrouff2022diagnosing}
Jessica Schrouff, Natalie Harris, Sanmi Koyejo, Ibrahim~M Alabdulmohsin, Eva Schnider, Krista Opsahl-Ong, Alexander Brown, Subhrajit Roy, Diana Mincu, Christina Chen, et~al.,
\newblock ``Diagnosing failures of fairness transfer across distribution shift in real-world medical settings,''
\newblock {\em Advances in Neural Information Processing Systems}, vol. 35, pp. 19304--19318, 2022.

\bibitem{lee2025using}
Simon~A Lee, Helio Halperin, Yanai Halperin, Trevor Brokowski, and Jeffrey~N Chiang,
\newblock ``Using foundation models to prescribe patients proper antibiotics,''
\newblock in {\em AAAI Bridge Program on AI for Medicine and Healthcare}. PMLR, 2025, pp. 121--132.

\bibitem{steinberg2023motor}
Ethan Steinberg, Jason Fries, Yizhe Xu, and Nigam Shah,
\newblock ``Motor: a time-to-event foundation model for structured medical records,''
\newblock {\em arXiv preprint arXiv:2301.03150}, 2023.

\bibitem{steinberg2024motor}
Ethan Steinberg, Yizhe Xu, Jason~Alan Fries, and Nigam Shah,
\newblock ``{MOTOR}: A time-to-event foundation model for structured medical records,''
\newblock in {\em The Twelfth International Conference on Learning Representations}, 2024.

\bibitem{wornowcontext}
Michael Wornow, Suhana Bedi, Miguel Angel~Fuentes Hernandez, Ethan Steinberg, Jason~Alan Fries, Christopher Re, Sanmi Koyejo, and Nigam Shah,
\newblock ``Context clues: Evaluating long context models for clinical prediction tasks on ehr data,''
\newblock in {\em The Thirteenth International Conference on Learning Representations}, 2025.

\bibitem{wornow2024context}
Michael Wornow, Suhana Bedi, Miguel Angel~Fuentes Hernandez, Ethan Steinberg, Jason~Alan Fries, Christopher R{\'e}, Sanmi Koyejo, and Nigam~H Shah,
\newblock ``Context clues: Evaluating long context models for clinical prediction tasks on ehrs,''
\newblock {\em arXiv preprint arXiv:2412.16178}, 2024.

\bibitem{lee2025clinical}
Simon~A Lee, Sujay Jain, Alex Chen, Kyoka Ono, Arabdha Biswas, {\'A}kos Rudas, Jennifer Fang, and Jeffrey~N Chiang,
\newblock ``Clinical decision support using pseudo-notes from multiple streams of ehr data,''
\newblock {\em npj Digital Medicine}, vol. 8, no. 1, pp. 394, 2025.

\bibitem{sellergren2025medgemmatechnicalreport}
Andrew Sellergren, Sahar Kazemzadeh, Tiam Jaroensri, Atilla Kiraly, Madeleine Traverse, Timo Kohlberger, Shawn Xu, Fayaz Jamil, Cían Hughes, Charles Lau, Justin Chen, Fereshteh Mahvar, Liron Yatziv, Tiffany Chen, Bram Sterling, Stefanie~Anna Baby, Susanna~Maria Baby, Jeremy Lai, Samuel Schmidgall, Lu~Yang, Kejia Chen, Per Bjornsson, Shashir Reddy, Ryan Brush, Kenneth Philbrick, Mercy Asiedu, Ines Mezerreg, Howard Hu, Howard Yang, Richa Tiwari, Sunny Jansen, Preeti Singh, Yun Liu, Shekoofeh Azizi, Aishwarya Kamath, Johan Ferret, Shreya Pathak, Nino Vieillard, Ramona Merhej, Sarah Perrin, Tatiana Matejovicova, Alexandre Ramé, Morgane Riviere, Louis Rouillard, Thomas Mesnard, Geoffrey Cideron, Jean bastien Grill, Sabela Ramos, Edouard Yvinec, Michelle Casbon, Elena Buchatskaya, Jean-Baptiste Alayrac, Dmitry Lepikhin, Vlad Feinberg, Sebastian Borgeaud, Alek Andreev, Cassidy Hardin, Robert Dadashi, Léonard Hussenot, Armand Joulin, Olivier Bachem, Yossi Matias, Katherine Chou, Avinatan Hassidim, Kavi Goel,
  Clement Farabet, Joelle Barral, Tris Warkentin, Jonathon Shlens, David Fleet, Victor Cotruta, Omar Sanseviero, Gus Martins, Phoebe Kirk, Anand Rao, Shravya Shetty, David~F. Steiner, Can Kirmizibayrak, Rory Pilgrim, Daniel Golden, and Lin Yang,
\newblock ``Medgemma technical report,'' 2025.

\bibitem{an2025dk}
Ulzee An, Simon~A Lee, Moonseong Jeong, Aditya Gorla, Jeffrey~N Chiang, and Sriram Sankararaman,
\newblock ``Dk-behrt: Teaching language models international classification of disease (icd) codes using known disease descriptions,''
\newblock in {\em AAAI Bridge Program on AI for Medicine and Healthcare}. PMLR, 2025, pp. 133--143.

\bibitem{wornow2023shaky}
Michael Wornow, Yizhe Xu, Rahul Thapa, Birju Patel, Ethan Steinberg, Scott Fleming, Michael~A Pfeffer, Jason Fries, and Nigam~H Shah,
\newblock ``The shaky foundations of large language models and foundation models for electronic health records,''
\newblock {\em npj digital medicine}, vol. 6, no. 1, pp. 135, 2023.

\bibitem{steinberg2021language}
Ethan Steinberg, Ken Jung, Jason~A Fries, Conor~K Corbin, Stephen~R Pfohl, and Nigam~H Shah,
\newblock ``Language models are an effective representation learning technique for electronic health record data,''
\newblock {\em Journal of biomedical informatics}, vol. 113, pp. 103637, 2021.

\bibitem{odgaard2024core}
Mikkel Odgaard, Kiril~Vadimovic Klein, Sanne~M{\o}ller Thysen, Espen Jimenez-Solem, Martin Sillesen, and Mads Nielsen,
\newblock ``Core-behrt: A carefully optimized and rigorously evaluated behrt,''
\newblock {\em arXiv preprint arXiv:2404.15201}, 2024.

\bibitem{cui2025timer}
Hejie Cui, Alyssa Unell, Bowen Chen, Jason~Alan Fries, Emily Alsentzer, Sanmi Koyejo, and Nigam~H Shah,
\newblock ``Timer: Temporal instruction modeling and evaluation for longitudinal clinical records,''
\newblock {\em NPJ Digital Medicine}, vol. 8, no. 1, pp. 577, 2025.

\bibitem{shmatko2025learning}
Artem Shmatko, Alexander~Wolfgang Jung, Kumar Gaurav, S{\o}ren Brunak, Laust~Hvas Mortensen, Ewan Birney, Tom Fitzgerald, and Moritz Gerstung,
\newblock ``Learning the natural history of human disease with generative transformers,''
\newblock {\em Nature}, vol. 647, no. 8088, pp. 248--256, 2025.

\bibitem{wornow2024ehrshot}
Michael Wornow, Rahul Thapa, Ethan Steinberg, Jason Fries, and Nigam Shah,
\newblock ``Ehrshot: An ehr benchmark for few-shot evaluation of foundation models,''
\newblock {\em Advances in Neural Information Processing Systems}, vol. 36, 2024.

\bibitem{openai2024gpt4technicalreport}
OpenAI, Josh Achiam, Steven Adler, Sandhini Agarwal, Lama Ahmad, Ilge Akkaya, Florencia~Leoni Aleman, Diogo Almeida, Janko Altenschmidt, Sam Altman, Shyamal Anadkat, Red Avila, Igor Babuschkin, Suchir Balaji, Valerie Balcom, Paul Baltescu, Haiming Bao, Mohammad Bavarian, Jeff Belgum, Irwan Bello, Jake Berdine, Gabriel Bernadett-Shapiro, Christopher Berner, Lenny Bogdonoff, Oleg Boiko, Madelaine Boyd, Anna-Luisa Brakman, Greg Brockman, Tim Brooks, Miles Brundage, Kevin Button, Trevor Cai, Rosie Campbell, Andrew Cann, Brittany Carey, Chelsea Carlson, Rory Carmichael, Brooke Chan, Che Chang, Fotis Chantzis, Derek Chen, Sully Chen, Ruby Chen, Jason Chen, Mark Chen, Ben Chess, Chester Cho, Casey Chu, Hyung~Won Chung, Dave Cummings, Jeremiah Currier, Yunxing Dai, Cory Decareaux, Thomas Degry, Noah Deutsch, Damien Deville, Arka Dhar, David Dohan, Steve Dowling, Sheila Dunning, Adrien Ecoffet, Atty Eleti, Tyna Eloundou, David Farhi, Liam Fedus, Niko Felix, Simón~Posada Fishman, Juston Forte, Isabella Fulford, Leo
  Gao, Elie Georges, Christian Gibson, Vik Goel, Tarun Gogineni, Gabriel Goh, Rapha Gontijo-Lopes, Jonathan Gordon, Morgan Grafstein, Scott Gray, Ryan Greene, Joshua Gross, Shixiang~Shane Gu, Yufei Guo, Chris Hallacy, Jesse Han, Jeff Harris, Yuchen He, Mike Heaton, Johannes Heidecke, Chris Hesse, Alan Hickey, Wade Hickey, Peter Hoeschele, Brandon Houghton, Kenny Hsu, Shengli Hu, Xin Hu, Joost Huizinga, Shantanu Jain, Shawn Jain, Joanne Jang, Angela Jiang, Roger Jiang, Haozhun Jin, Denny Jin, Shino Jomoto, Billie Jonn, Heewoo Jun, Tomer Kaftan, Łukasz Kaiser, Ali Kamali, Ingmar Kanitscheider, Nitish~Shirish Keskar, Tabarak Khan, Logan Kilpatrick, Jong~Wook Kim, Christina Kim, Yongjik Kim, Jan~Hendrik Kirchner, Jamie Kiros, Matt Knight, Daniel Kokotajlo, Łukasz Kondraciuk, Andrew Kondrich, Aris Konstantinidis, Kyle Kosic, Gretchen Krueger, Vishal Kuo, Michael Lampe, Ikai Lan, Teddy Lee, Jan Leike, Jade Leung, Daniel Levy, Chak~Ming Li, Rachel Lim, Molly Lin, Stephanie Lin, Mateusz Litwin, Theresa Lopez, Ryan
  Lowe, Patricia Lue, Anna Makanju, Kim Malfacini, Sam Manning, Todor Markov, Yaniv Markovski, Bianca Martin, Katie Mayer, Andrew Mayne, Bob McGrew, Scott~Mayer McKinney, Christine McLeavey, Paul McMillan, Jake McNeil, David Medina, Aalok Mehta, Jacob Menick, Luke Metz, Andrey Mishchenko, Pamela Mishkin, Vinnie Monaco, Evan Morikawa, Daniel Mossing, Tong Mu, Mira Murati, Oleg Murk, David Mély, Ashvin Nair, Reiichiro Nakano, Rajeev Nayak, Arvind Neelakantan, Richard Ngo, Hyeonwoo Noh, Long Ouyang, Cullen O'Keefe, Jakub Pachocki, Alex Paino, Joe Palermo, Ashley Pantuliano, Giambattista Parascandolo, Joel Parish, Emy Parparita, Alex Passos, Mikhail Pavlov, Andrew Peng, Adam Perelman, Filipe de~Avila Belbute~Peres, Michael Petrov, Henrique~Ponde de~Oliveira~Pinto, Michael, Pokorny, Michelle Pokrass, Vitchyr~H. Pong, Tolly Powell, Alethea Power, Boris Power, Elizabeth Proehl, Raul Puri, Alec Radford, Jack Rae, Aditya Ramesh, Cameron Raymond, Francis Real, Kendra Rimbach, Carl Ross, Bob Rotsted, Henri Roussez,
  Nick Ryder, Mario Saltarelli, Ted Sanders, Shibani Santurkar, Girish Sastry, Heather Schmidt, David Schnurr, John Schulman, Daniel Selsam, Kyla Sheppard, Toki Sherbakov, Jessica Shieh, Sarah Shoker, Pranav Shyam, Szymon Sidor, Eric Sigler, Maddie Simens, Jordan Sitkin, Katarina Slama, Ian Sohl, Benjamin Sokolowsky, Yang Song, Natalie Staudacher, Felipe~Petroski Such, Natalie Summers, Ilya Sutskever, Jie Tang, Nikolas Tezak, Madeleine~B. Thompson, Phil Tillet, Amin Tootoonchian, Elizabeth Tseng, Preston Tuggle, Nick Turley, Jerry Tworek, Juan Felipe~Cerón Uribe, Andrea Vallone, Arun Vijayvergiya, Chelsea Voss, Carroll Wainwright, Justin~Jay Wang, Alvin Wang, Ben Wang, Jonathan Ward, Jason Wei, CJ~Weinmann, Akila Welihinda, Peter Welinder, Jiayi Weng, Lilian Weng, Matt Wiethoff, Dave Willner, Clemens Winter, Samuel Wolrich, Hannah Wong, Lauren Workman, Sherwin Wu, Jeff Wu, Michael Wu, Kai Xiao, Tao Xu, Sarah Yoo, Kevin Yu, Qiming Yuan, Wojciech Zaremba, Rowan Zellers, Chong Zhang, Marvin Zhang, Shengjia
  Zhao, Tianhao Zheng, Juntang Zhuang, William Zhuk, and Barret Zoph,
\newblock ``Gpt-4 technical report,'' 2024.

\bibitem{fallahpour2024ehrmamba}
Adibvafa Fallahpour, Mahshid Alinoori, Wenqian Ye, Xu~Cao, Arash Afkanpour, and Amrit Krishnan,
\newblock ``Ehrmamba: Towards generalizable and scalable foundation models for electronic health records,''
\newblock {\em arXiv preprint arXiv:2405.14567}, 2024.

\bibitem{lee2025towards}
Simon~A Lee, Cyrus Tanade, Hao Zhou, Juhyeon Lee, Megha Thukral, Baiying Lu, and Sharanya~Arcot Desai,
\newblock ``Towards on-device foundation models for raw wearable signals,''
\newblock in {\em NeurIPS 2025 Workshop on Learning from Time Series for Health}, 2025.

\bibitem{hollmann2025accurate}
Noah Hollmann, Samuel M{\"u}ller, Lennart Purucker, Arjun Krishnakumar, Max K{\"o}rfer, Shi~Bin Hoo, Robin~Tibor Schirrmeister, and Frank Hutter,
\newblock ``Accurate predictions on small data with a tabular foundation model,''
\newblock {\em Nature}, vol. 637, no. 8045, pp. 319--326, 2025.

\bibitem{lee2024can}
Simon~A Lee and Timothy Lindsey,
\newblock ``Can large language models abstract medical coded language?,''
\newblock {\em arXiv preprint arXiv:2403.10822}, 2024.

\bibitem{jin2023time}
Ming Jin, Shiyu Wang, Lintao Ma, Zhixuan Chu, James~Y Zhang, Xiaoming Shi, Pin-Yu Chen, Yuxuan Liang, Yuan-Fang Li, Shirui Pan, et~al.,
\newblock ``Time-llm: Time series forecasting by reprogramming large language models,''
\newblock {\em arXiv preprint arXiv:2310.01728}, 2023.

\bibitem{chang2025llm4ts}
Ching Chang, Wei-Yao Wang, Wen-Chih Peng, and Tien-Fu Chen,
\newblock ``Llm4ts: Aligning pre-trained llms as data-efficient time-series forecasters,''
\newblock {\em ACM Transactions on Intelligent Systems and Technology}, vol. 16, no. 3, pp. 1--20, 2025.

\bibitem{belyaeva2023multimodal}
Anastasiya Belyaeva, Justin Cosentino, Farhad Hormozdiari, Krish Eswaran, Shravya Shetty, Greg Corrado, Andrew Carroll, Cory~Y McLean, and Nicholas~A Furlotte,
\newblock ``Multimodal llms for health grounded in individual-specific data,''
\newblock in {\em Workshop on Machine Learning for Multimodal Healthcare Data}. Springer, 2023, pp. 86--102.

\bibitem{karabacak2023embracing}
Mert Karabacak and Konstantinos Margetis,
\newblock ``Embracing large language models for medical applications: Opportunities and challenges,''
\newblock {\em Cureus}, vol. 15, no. 5, 2023.

\bibitem{van2023clinical}
Dave Van~Veen, Cara Van~Uden, Louis Blankemeier, Jean-Benoit Delbrouck, Asad Aali, Christian Bluethgen, Anuj Pareek, Malgorzata Polacin, Eduardo~Pontes Reis, Anna Seehofnerova, et~al.,
\newblock ``Clinical text summarization: Adapting large language models can outperform human experts,''
\newblock {\em Research Square}, 2023.

\bibitem{ono2024text}
Kyoka Ono and Simon~A Lee,
\newblock ``Text serialization and their relationship with the conventional paradigms of tabular machine learning,''
\newblock {\em arXiv preprint arXiv:2406.13846}, 2024.

\bibitem{grattafiori2024llama3herdmodels}
Aaron Grattafiori, Abhimanyu Dubey, Abhinav Jauhri, Abhinav Pandey, Abhishek Kadian, Ahmad Al-Dahle, Aiesha Letman, Akhil Mathur, Alan Schelten, Alex Vaughan, Amy Yang, Angela Fan, Anirudh Goyal, Anthony Hartshorn, Aobo Yang, Archi Mitra, Archie Sravankumar, Artem Korenev, Arthur Hinsvark, Arun Rao, Aston Zhang, Aurelien Rodriguez, Austen Gregerson, Ava Spataru, Baptiste Roziere, Bethany Biron, Binh Tang, Bobbie Chern, Charlotte Caucheteux, Chaya Nayak, Chloe Bi, Chris Marra, Chris McConnell, Christian Keller, Christophe Touret, Chunyang Wu, Corinne Wong, Cristian~Canton Ferrer, Cyrus Nikolaidis, Damien Allonsius, Daniel Song, Danielle Pintz, Danny Livshits, Danny Wyatt, David Esiobu, Dhruv Choudhary, Dhruv Mahajan, Diego Garcia-Olano, Diego Perino, Dieuwke Hupkes, Egor Lakomkin, Ehab AlBadawy, Elina Lobanova, Emily Dinan, Eric~Michael Smith, Filip Radenovic, Francisco Guzmán, Frank Zhang, Gabriel Synnaeve, Gabrielle Lee, Georgia~Lewis Anderson, Govind Thattai, Graeme Nail, Gregoire Mialon, Guan Pang,
  Guillem Cucurell, Hailey Nguyen, Hannah Korevaar, Hu~Xu, Hugo Touvron, Iliyan Zarov, Imanol~Arrieta Ibarra, Isabel Kloumann, Ishan Misra, Ivan Evtimov, Jack Zhang, Jade Copet, Jaewon Lee, Jan Geffert, Jana Vranes, Jason Park, Jay Mahadeokar, Jeet Shah, Jelmer van~der Linde, Jennifer Billock, Jenny Hong, Jenya Lee, Jeremy Fu, Jianfeng Chi, Jianyu Huang, Jiawen Liu, Jie Wang, Jiecao Yu, Joanna Bitton, Joe Spisak, Jongsoo Park, Joseph Rocca, Joshua Johnstun, Joshua Saxe, Junteng Jia, Kalyan~Vasuden Alwala, Karthik Prasad, Kartikeya Upasani, Kate Plawiak, Ke~Li, Kenneth Heafield, Kevin Stone, Khalid El-Arini, Krithika Iyer, Kshitiz Malik, Kuenley Chiu, Kunal Bhalla, Kushal Lakhotia, Lauren Rantala-Yeary, Laurens van~der Maaten, Lawrence Chen, Liang Tan, Liz Jenkins, Louis Martin, Lovish Madaan, Lubo Malo, Lukas Blecher, Lukas Landzaat, Luke de~Oliveira, Madeline Muzzi, Mahesh Pasupuleti, Mannat Singh, Manohar Paluri, Marcin Kardas, Maria Tsimpoukelli, Mathew Oldham, Mathieu Rita, Maya Pavlova, Melanie Kambadur,
  Mike Lewis, Min Si, Mitesh~Kumar Singh, Mona Hassan, Naman Goyal, Narjes Torabi, Nikolay Bashlykov, Nikolay Bogoychev, Niladri Chatterji, Ning Zhang, Olivier Duchenne, Onur Çelebi, Patrick Alrassy, Pengchuan Zhang, Pengwei Li, Petar Vasic, Peter Weng, Prajjwal Bhargava, Pratik Dubal, Praveen Krishnan, Punit~Singh Koura, Puxin Xu, Qing He, Qingxiao Dong, Ragavan Srinivasan, Raj Ganapathy, Ramon Calderer, Ricardo~Silveira Cabral, Robert Stojnic, Roberta Raileanu, Rohan Maheswari, Rohit Girdhar, Rohit Patel, Romain Sauvestre, Ronnie Polidoro, Roshan Sumbaly, Ross Taylor, Ruan Silva, Rui Hou, Rui Wang, Saghar Hosseini, Sahana Chennabasappa, Sanjay Singh, Sean Bell, Seohyun~Sonia Kim, Sergey Edunov, Shaoliang Nie, Sharan Narang, Sharath Raparthy, Sheng Shen, Shengye Wan, Shruti Bhosale, Shun Zhang, Simon Vandenhende, Soumya Batra, Spencer Whitman, Sten Sootla, Stephane Collot, Suchin Gururangan, Sydney Borodinsky, Tamar Herman, Tara Fowler, Tarek Sheasha, Thomas Georgiou, Thomas Scialom, Tobias Speckbacher,
  Todor Mihaylov, Tong Xiao, Ujjwal Karn, Vedanuj Goswami, Vibhor Gupta, Vignesh Ramanathan, Viktor Kerkez, Vincent Gonguet, Virginie Do, Vish Vogeti, Vítor Albiero, Vladan Petrovic, Weiwei Chu, Wenhan Xiong, Wenyin Fu, Whitney Meers, Xavier Martinet, Xiaodong Wang, Xiaofang Wang, Xiaoqing~Ellen Tan, Xide Xia, Xinfeng Xie, Xuchao Jia, Xuewei Wang, Yaelle Goldschlag, Yashesh Gaur, Yasmine Babaei, Yi~Wen, Yiwen Song, Yuchen Zhang, Yue Li, Yuning Mao, Zacharie~Delpierre Coudert, Zheng Yan, Zhengxing Chen, Zoe Papakipos, Aaditya Singh, Aayushi Srivastava, Abha Jain, Adam Kelsey, Adam Shajnfeld, Adithya Gangidi, Adolfo Victoria, Ahuva Goldstand, Ajay Menon, Ajay Sharma, Alex Boesenberg, Alexei Baevski, Allie Feinstein, Amanda Kallet, Amit Sangani, Amos Teo, Anam Yunus, Andrei Lupu, Andres Alvarado, Andrew Caples, Andrew Gu, Andrew Ho, Andrew Poulton, Andrew Ryan, Ankit Ramchandani, Annie Dong, Annie Franco, Anuj Goyal, Aparajita Saraf, Arkabandhu Chowdhury, Ashley Gabriel, Ashwin Bharambe, Assaf Eisenman, Azadeh
  Yazdan, Beau James, Ben Maurer, Benjamin Leonhardi, Bernie Huang, Beth Loyd, Beto~De Paola, Bhargavi Paranjape, Bing Liu, Bo~Wu, Boyu Ni, Braden Hancock, Bram Wasti, Brandon Spence, Brani Stojkovic, Brian Gamido, Britt Montalvo, Carl Parker, Carly Burton, Catalina Mejia, Ce~Liu, Changhan Wang, Changkyu Kim, Chao Zhou, Chester Hu, Ching-Hsiang Chu, Chris Cai, Chris Tindal, Christoph Feichtenhofer, Cynthia Gao, Damon Civin, Dana Beaty, Daniel Kreymer, Daniel Li, David Adkins, David Xu, Davide Testuggine, Delia David, Devi Parikh, Diana Liskovich, Didem Foss, Dingkang Wang, Duc Le, Dustin Holland, Edward Dowling, Eissa Jamil, Elaine Montgomery, Eleonora Presani, Emily Hahn, Emily Wood, Eric-Tuan Le, Erik Brinkman, Esteban Arcaute, Evan Dunbar, Evan Smothers, Fei Sun, Felix Kreuk, Feng Tian, Filippos Kokkinos, Firat Ozgenel, Francesco Caggioni, Frank Kanayet, Frank Seide, Gabriela~Medina Florez, Gabriella Schwarz, Gada Badeer, Georgia Swee, Gil Halpern, Grant Herman, Grigory Sizov, Guangyi, Zhang, Guna
  Lakshminarayanan, Hakan Inan, Hamid Shojanazeri, Han Zou, Hannah Wang, Hanwen Zha, Haroun Habeeb, Harrison Rudolph, Helen Suk, Henry Aspegren, Hunter Goldman, Hongyuan Zhan, Ibrahim Damlaj, Igor Molybog, Igor Tufanov, Ilias Leontiadis, Irina-Elena Veliche, Itai Gat, Jake Weissman, James Geboski, James Kohli, Janice Lam, Japhet Asher, Jean-Baptiste Gaya, Jeff Marcus, Jeff Tang, Jennifer Chan, Jenny Zhen, Jeremy Reizenstein, Jeremy Teboul, Jessica Zhong, Jian Jin, Jingyi Yang, Joe Cummings, Jon Carvill, Jon Shepard, Jonathan McPhie, Jonathan Torres, Josh Ginsburg, Junjie Wang, Kai Wu, Kam~Hou U, Karan Saxena, Kartikay Khandelwal, Katayoun Zand, Kathy Matosich, Kaushik Veeraraghavan, Kelly Michelena, Keqian Li, Kiran Jagadeesh, Kun Huang, Kunal Chawla, Kyle Huang, Lailin Chen, Lakshya Garg, Lavender A, Leandro Silva, Lee Bell, Lei Zhang, Liangpeng Guo, Licheng Yu, Liron Moshkovich, Luca Wehrstedt, Madian Khabsa, Manav Avalani, Manish Bhatt, Martynas Mankus, Matan Hasson, Matthew Lennie, Matthias Reso, Maxim
  Groshev, Maxim Naumov, Maya Lathi, Meghan Keneally, Miao Liu, Michael~L. Seltzer, Michal Valko, Michelle Restrepo, Mihir Patel, Mik Vyatskov, Mikayel Samvelyan, Mike Clark, Mike Macey, Mike Wang, Miquel~Jubert Hermoso, Mo~Metanat, Mohammad Rastegari, Munish Bansal, Nandhini Santhanam, Natascha Parks, Natasha White, Navyata Bawa, Nayan Singhal, Nick Egebo, Nicolas Usunier, Nikhil Mehta, Nikolay~Pavlovich Laptev, Ning Dong, Norman Cheng, Oleg Chernoguz, Olivia Hart, Omkar Salpekar, Ozlem Kalinli, Parkin Kent, Parth Parekh, Paul Saab, Pavan Balaji, Pedro Rittner, Philip Bontrager, Pierre Roux, Piotr Dollar, Polina Zvyagina, Prashant Ratanchandani, Pritish Yuvraj, Qian Liang, Rachad Alao, Rachel Rodriguez, Rafi Ayub, Raghotham Murthy, Raghu Nayani, Rahul Mitra, Rangaprabhu Parthasarathy, Raymond Li, Rebekkah Hogan, Robin Battey, Rocky Wang, Russ Howes, Ruty Rinott, Sachin Mehta, Sachin Siby, Sai~Jayesh Bondu, Samyak Datta, Sara Chugh, Sara Hunt, Sargun Dhillon, Sasha Sidorov, Satadru Pan, Saurabh Mahajan,
  Saurabh Verma, Seiji Yamamoto, Sharadh Ramaswamy, Shaun Lindsay, Shaun Lindsay, Sheng Feng, Shenghao Lin, Shengxin~Cindy Zha, Shishir Patil, Shiva Shankar, Shuqiang Zhang, Shuqiang Zhang, Sinong Wang, Sneha Agarwal, Soji Sajuyigbe, Soumith Chintala, Stephanie Max, Stephen Chen, Steve Kehoe, Steve Satterfield, Sudarshan Govindaprasad, Sumit Gupta, Summer Deng, Sungmin Cho, Sunny Virk, Suraj Subramanian, Sy~Choudhury, Sydney Goldman, Tal Remez, Tamar Glaser, Tamara Best, Thilo Koehler, Thomas Robinson, Tianhe Li, Tianjun Zhang, Tim Matthews, Timothy Chou, Tzook Shaked, Varun Vontimitta, Victoria Ajayi, Victoria Montanez, Vijai Mohan, Vinay~Satish Kumar, Vishal Mangla, Vlad Ionescu, Vlad Poenaru, Vlad~Tiberiu Mihailescu, Vladimir Ivanov, Wei Li, Wenchen Wang, Wenwen Jiang, Wes Bouaziz, Will Constable, Xiaocheng Tang, Xiaojian Wu, Xiaolan Wang, Xilun Wu, Xinbo Gao, Yaniv Kleinman, Yanjun Chen, Ye~Hu, Ye~Jia, Ye~Qi, Yenda Li, Yilin Zhang, Ying Zhang, Yossi Adi, Youngjin Nam, Yu, Wang, Yu~Zhao, Yuchen Hao, Yundi
  Qian, Yunlu Li, Yuzi He, Zach Rait, Zachary DeVito, Zef Rosnbrick, Zhaoduo Wen, Zhenyu Yang, Zhiwei Zhao, and Zhiyu Ma,
\newblock ``The llama 3 herd of models,'' 2024.

\bibitem{lin2025case}
Yihan Lin, Zhirong~Bella Yu, and Simon Lee,
\newblock ``A case study exploring the current landscape of synthetic medical record generation with commercial llms,''
\newblock {\em arXiv preprint arXiv:2504.14657}, 2025.

\bibitem{mumtaz2023llms}
Ummara Mumtaz, Awais Ahmed, and Summaya Mumtaz,
\newblock ``Llms-healthcare: Current applications and challenges of large language models in various medical specialties,''
\newblock {\em arXiv preprint arXiv:2311.12882}, 2023.

\end{thebibliography}

\appendix
\section{Appendix}

\subsection{Spectral Properties and Effective Rank}

We provide additional theoretical context for the spectral objectives used in dense feature learning. Let $\Sigma_Z = \frac{1}{N} Z^\top Z$ denote the empirical covariance matrix of learned representations, with eigenvalues $\lambda_1 \geq \lambda_2 \geq \dots \geq \lambda_d \geq 0$. While the algebraic rank of $\Sigma_Z$ captures exact degeneracy, it is often too brittle to reflect the usable dimensionality of a representation. We therefore rely on the notion of effective rank, defined as
\[
\mathrm{rank}_{\mathrm{eff}}(\Sigma_Z) 
= 
\exp\left( - \sum_{i=1}^d p_i \log p_i \right),
\quad
p_i = \frac{\lambda_i}{\sum_j \lambda_j}.
\]
This quantity corresponds to the exponential of the Shannon entropy of the normalized spectrum and provides a smooth measure of dimensional utilization. Maximizing effective rank discourages concentration of variance while remaining invariant to isotropic rescaling.

Our spectral spreading loss can be interpreted as minimizing the squared distance between the normalized spectrum of $\Sigma_Z$ and the uniform distribution over $d$ dimensions. In particular, letting $\tilde{\Sigma}_Z = \Sigma_Z / \mathrm{tr}(\Sigma_Z)$, we have
\[
\mathcal{L}_{\text{spec}} 
= 
\left\| \tilde{\Sigma}_Z - \frac{1}{d} I \right\|_F^2
=
\sum_{i=1}^d \left( \tilde{\lambda}_i - \frac{1}{d} \right)^2,
\]
where $\tilde{\lambda}_i$ are the eigenvalues of $\tilde{\Sigma}_Z$. This shows that the loss penalizes deviations from equal spectral mass, directly controlling anisotropy.

\subsection{Subspace Geometry and Projection Metrics}

We elaborate on the subspace consistency objective and its geometric interpretation. Given two $k$-dimensional subspaces $\mathcal{U}, \mathcal{V} \subset \mathbb{R}^d$ with orthonormal bases $U, V \in \mathbb{R}^{d \times k}$, the projection matrices $P_U = UU^\top$ and $P_V = VV^\top$ uniquely represent these subspaces. The Frobenius distance between projections satisfies
\[
\| P_U - P_V \|_F^2 = 2k - 2 \| U^\top V \|_F^2,
\]
which is equal to twice the sum of squared sines of the principal angles between $\mathcal{U}$ and $\mathcal{V}$. Consequently, minimizing this distance encourages alignment of dominant directions while remaining invariant to basis choice within each subspace.

In longitudinal medical data, this formulation naturally captures the intuition that latent modes of variation should persist over time, even as the specific embedding of each observation changes. By operating on subspaces rather than individual points, the loss avoids over-constraining representations and allows flexibility in local geometry.

\subsection{Orthogonality, Conditioning, and Downstream Linear Models}

The batch-wise orthogonality objective can be viewed as a soft conditioning constraint. For a normalized batch embedding matrix $Z_B$, the Gram matrix $G_B = \frac{1}{B} Z_B^\top Z_B$ approximates the identity when features are decorrelated. Deviations from identity indicate redundancy or imbalance across dimensions. Penalizing $\|G_B - I\|_F^2$ therefore improves the numerical conditioning of linear probes trained on top of frozen representations.

Well-conditioned representations reduce sensitivity to estimation noise in downstream models. In particular, for linear regression or classification, the variance of estimated coefficients depends inversely on the smallest eigenvalues of $\Sigma_Z$. By preventing spectral collapse, dense feature learning stabilizes downstream optimization and improves sample efficiency, a critical consideration in medical settings with limited labeled data.

\subsection{Optimization Considerations}

Although the objectives involve covariance matrices and singular vectors, all terms are differentiable with respect to $Z$ and hence to encoder parameters $\theta$. In practice, we compute covariance statistics over mini-batches and approximate leading singular vectors using power iteration. Empirically, we observe stable training dynamics without the need for explicit whitening or matrix inversion.

We emphasize that the method does not require exact satisfaction of linear constraints. Each term acts as a soft regularizer, allowing the representation to deviate from idealized structure when supported by the data. This flexibility is essential for heterogeneous medical datasets, where strict linear assumptions rarely hold globally.

\subsection{Interpretation as Basis Learning}

Finally, we note that dense feature learning can be interpreted as learning a data-adaptive basis for medical observations. From this perspective, the encoder constructs a coordinate system in which clinically relevant variation is spread across axes, stable across contexts, and minimally redundant. Downstream tasks then correspond to selecting directions or subspaces within this basis. This view highlights the separation between representation learning and task learning and suggests that many clinical prediction problems can be addressed through linear operations once an appropriate basis has been learned.

\end{document}